\pdfoutput=1
\documentclass[11pt]{article}

\usepackage[final]{acl}

\usepackage{times}
\usepackage{latexsym}

\usepackage[T1]{fontenc}
\usepackage[utf8]{inputenc}

\usepackage{microtype}

\usepackage{inconsolata}

\usepackage{graphicx}
\usepackage{multirow}
\usepackage{multicol}
\usepackage{booktabs}
\usepackage{tipa}
\usepackage{amsmath}
\usepackage{amssymb}
\usepackage{xcolor}
\usepackage{colortbl}

\definecolor{LightBlue}{rgb}{0.8235,0.9737,0.9882}
\usepackage{tablefootnote,footnotehyper}

\title{Zero-Shot Cross-Lingual NER Using Phonemic Representations for Low-Resource Languages}

\author{
 \textbf{Jimin Sohn$^*$\textsuperscript{1}},
 \textbf{Haeji Jung$^*$\textsuperscript{2}},
 \textbf{Alex Cheng\textsuperscript{3}},
 \textbf{Jooeon Kang\textsuperscript{4}},
 \textbf{Yilin Du\textsuperscript{3}},
 \textbf{David R. Mortensen\textsuperscript{3}}
\\
\\
 \textsuperscript{1}GIST, South Korea,
 \textsuperscript{2}Korea University, South Korea, \\
 \textsuperscript{3}Carnegie Mellon University, USA,
 \textsuperscript{4}Sogang University, South Korea
\\
\small{
   \href{mailto:estelle26598@gm.gist.ac.kr}{estelle26598@gm.gist.ac.kr}, \href{mailto:gpwl0709@korea.ac.kr}{gpwl0709@korea.ac.kr}
 }
}

\begin{document}
\maketitle
\def\thefootnote{*}\footnotetext{These authors contributed equally to this work.}\def\thefootnote{\arabic{footnote}}
\begin{abstract}
Existing zero-shot cross-lingual NER approaches require substantial prior knowledge of the target language, which is impractical for low-resource languages.
In this paper, we propose a novel approach to NER using phonemic representation based on the International Phonetic Alphabet (IPA) to bridge the gap between representations of different languages.
Our experiments show that our method significantly outperforms baseline models in extremely low-resource languages, with the highest average F1 score (46.38\%) and lowest standard deviation (12.67), particularly demonstrating its robustness with non-Latin scripts. Our
codes are available at \href{https://github.com/Gabriel819/zeroshot_ner.git}{https://github.com/Gabriel819/zeroshot\_ner.git}
\end{abstract}

\section{Introduction}
Named entity recognition (NER) plays a crucial role in many Natural Language Processing (NLP) tasks. Achieving high performance in NER generally requires extensive resources for both sequence labeling and gazetteer training \cite{10.1145/3015467}.
 However, access to training resources for many low-resource languages (LRLs) is very limited,  motivating zero-shot approaches to the task. While various strategies have been explored to enhance zero-shot NER performance across languages, they required either parallel data or unlabeled corpora in the target language, which is difficult and sometimes impossible to obtain. 
 
\begin{figure}[t!]
    \centering
    \includegraphics[width=0.48\textwidth]{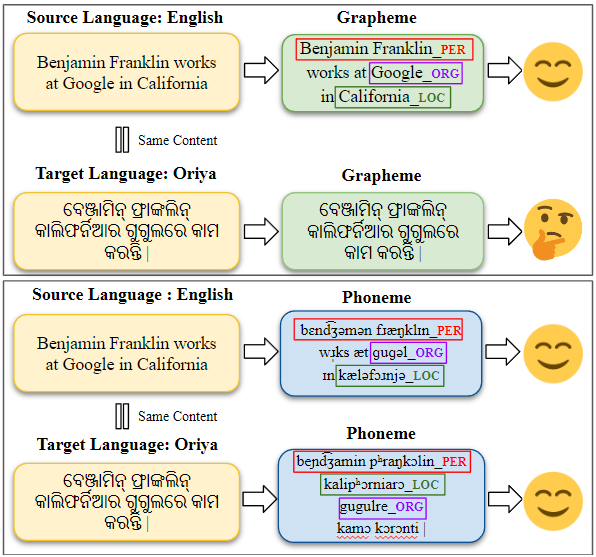}
    \caption{Zero-shot Cross-Lingual NER with IPA phonemes.}
    \label{fig:concept_fig}
\end{figure}

 Our work tackles zero-shot NER under a strict condition that disallows any target language training data. We decided to approach this condition by projecting data into an International Phonetic Alphabet (IPA) space. Since different languages often share similar pronunciations for the same entities, such as geopolitical entities and personal names (e.g., the word for China is /\textipa{\t{tS}ajn@}/ in English and /\textipa{\t{tS}ina}/ in Sinhala), the model trained on one language can be transferred to others without target-language training in NER.
 As shown in Figure \ref{fig:concept_fig}, we first convert orthographic scripts into IPA, and then fine-tune a pre-trained model on the phonemes of the source language, i.e., English. By using a shared notation system---IPA---we can apply the model to target languages directly. Our findings show that fine-tuning phoneme-based models outperforms traditional grapheme-based models (e.g., mBERT~\cite{devlin-etal-2019-bert}) by a large margin for LRLs not seen during pre-training. Furthermore, our approach demonstrates robustness with non-Latin scripts, exhibiting stable performance across languages with different writing systems.
 
\section{Related Work}
\subsection{Zero-shot Cross-lingual NER}
Recent approaches for zero-shot cross-lingual NER can be categorized into three groups based on how they use resources from target languages.
One line of work involves using translation between source and target languages to transfer NER capability~\cite{yang2022crop,liu-etal-2021-mulda,mo2024mclner}. These methods require parallel data from both languages, which is not always available. 
Alternatively, some methods use unlabeled target language data and adopt knowledge distillation without needing parallel data~\cite{deb-etal-2023-zero, li-etal-2022-unsupervised-multiple}.
However, these approaches are still not widely applicable to languages with extremely low-resources, as such languages often lack sufficient resources for training.
On the other hand,~\cite{rathore-etal-2023-zgul} assumes that no data in target language is available during training. While it provides a practical setting for extremely low-resource languages, it requires language adapters pre-trained on similar languages to the target language, as well as typological information (i.e., language family) of various languages.

We assume a very strict problem setting where the target language for zero-shot inference, as well as its typological information, is completely unavailable during training.
Unlike previous methods that rely on some of the target language data during training, we use IPA phonemes for NER, making our method entirely data-independent for the target language. It only relies on the availability of an easily constructed grapheme-to-phoneme (G2P) module.

\subsection{Phonemic Representation}
Phonological traits of languages are useful in understanding different languages, as they often share similar pronunciations for similar entities. 
It is particularly beneficial for NER, where many items, such as geopolitical entities and personal names, are pronounced similarly across various languages.
Moreover, phonemes are represented in IPA, which is shared across all languages. By providing a universal script, phonemic representations help the model better address low-resource languages, as the poor NER performance in these languages comes significantly from the model's limited ability to handle relevant scripts~\cite{muller2020being, severini2022towards}.

While phonological information has been shown to be helpful in language understanding for cross-lingual transfer~\cite{chaudhary-etal-2018-adapting,Sun2021AlternativeIS,bharadwaj-etal-2016-phonologically,leong2022phone}, few works have explored its benefits compared to orthographic input, particularly in zero-shot scenario where the target language is not available for fine-tuning.
Given that creating rule-based transcription module for most low-resource languages takes only a few hours and limited training, we use IPA to enable zero-shot cross-lingual NER on languages with very scarce resources, without requiring any additional corpus for those languages.

\section{Our Approach}
\subsection{NER with Phonemes}
In this paper, we conduct NER using phonetic transcriptions (IPA) instead of conventional orthographic text. Leveraging the standard practice of using multilingual pre-trained models for cross-lingual transfer, we employ XPhoneBERT~\cite{nguyen2023xphonebert}, a model pre-trained on phonemes from 94 different languages. By utilizing pre-trained phonemic representations, the model can fully utilize the phonological knowledge across diverse languages.

To create a phoneme-based version of the dataset originally containing graphemes, we convert the dataset into IPA representations. 
For G2P conversion of various languages, we use Epitran~\cite{Mortensen-et-al:2018} along with the CharsiuG2P toolkit~\cite{Zhu2022ByT5MF} which XPhoneBERT originally employed.
Epitran supports the transliteration of approximately 100 languages, including numerous low-resource languages. We apply transliteration at the word level, maintaining the pre-tokenized units consistent with the original version.

We adopt the BIO tagging scheme for entity tagging. As the phoneme is the input unit for the model, we assign each phoneme a named entity tag. Only the first phoneme segment of the first word of a named entity is assigned with a `\texttt{B}' tag, indicating the beginning of the entity. For example, the phoneme sequence ``\textipa{b\textepsilon n\t{dZ}@m@n} (Benjamin)'' comprises nine segments\footnote{Phoneme segmentation is performed using the Python library `segments,' as utilized in XPhoneBERT.}, and is labeled as \texttt{[``B-PER", ``I-PER", ...,``I-PER"]}.

\subsection{Cross-lingual Transfer to Unseen Languages}

We perform zero-shot named entity recognition on low-resource languages, where the model is only trained on a single high-resource language, in this case, English. Although the model is fine-tuned on a single language, its pre-training on approximately 100 languages allows it to retain some knowledge of other languages. We hypothesize that (i) each model will leverage its pre-trained knowledge on the target languages in performing NER, and (ii) phoneme-based models will generally achieve superior performance with unseen languages, benefiting from phonological traits shared across languages.

To investigate the generalizability of phonemic representations in extremely low-resource languages, we do not allow any access to the target language during training and exclude their typological information to keep our method language-agnostic. 
We use mBERT and CANINE as baselines, as these models are compatible with our problem setting, requiring no additional training data for the target languages.

\begin{table}[t]
\centering
\resizebox{\columnwidth}{!}{
\begin{tabular}{c|ccc|c|c}
\toprule
\multirow{2}{*}{Case} & \multicolumn{3}{c|}{Models} & \multirow{2}{*}{Languages} & \multirow{2}{*}{Num} \\
\cmidrule{2-4}
& M & C & X &&\\\midrule
1 & - & - & - & sin, som, mri, quy, uig, aii, kin, ilo & 8 \\
\midrule
2 & - & - & \checkmark & epo, khm, tuk, amh, mlt, ori, san, ina, grn, bel, kur, snd & 12 \\
\midrule
3 & \checkmark & \checkmark & - & tgk, yor, mar, jav, urd, msa, ceb, hrv, mal, tel, uzb, pan, kir & 13 \\
\bottomrule
\end{tabular}
}
\caption{\label{tab:cases_languages}
Languages for each case. M, C, X indicates mBERT, CANINE, and XPhoneBERT, respectively, and \checkmark represents the languages pre-trained on the model.
}
\end{table}
As shown in Table \ref{tab:cases_languages}, we define three sets of languages based on whether the language has been seen during pre-training of each model. Let $L$ be the set of all languages in our benchmark dataset that are able to be transliterated, $B$ the set of languages pre-trained on the baseline models, and $X$ the set of languages pre-trained on XPhoneBERT.\\ 
\textbf{Case 1}: ($L \setminus (B \cup X)$) includes languages not in the pre-training data for any models. \\
\textbf{Case 2}: ($(L \cap X) \setminus B$) includes languages in the pre-training data of XPhoneBERT only.\\
\textbf{Case 3}: ($(L \cap B) \setminus X$) includes languages in the pre-training data of mBERT and CANINE only.

\section{Experiments}

Here we provide information about the datasets and models we employed for the experiments. More  implementation details including hyperparameters are provided in Appendix \ref{sec:appendix-implementation_details}.

\subsection{Benchmark Dataset} 
We train and evaluate our method on the WikiANN NER datasets~\cite{pan-etal-2017-cross} which has three different named entity types: person (PER), organization (ORG), and location (LOC).
The models are trained only on English data and evaluated on various low-resource languages. We select languages that are (i) supported by either Epitran or CharsiuG2P toolkit for transliteration, and (ii) not included in the pre-training of at least one of the baseline models. This yields 33 languages in total, as listed in Table \ref{tab:cases_languages}.

\subsection{Baseline Models} 
We use mBERT~\cite{devlin-etal-2019-bert} and  CANINE~\cite{clark2022canine}, both grapheme-based language models, as baselines to compare to XPhoneBERT~\cite{nguyen2023xphonebert}, a phoneme-based language model.
All three models are BERT-like transformer architectures pre-trained on a Wikipedia corpora of multiple languages: mBERT and CANINE are trained on the same 104 languages, while XPhoneBERT is trained on 94 languages and locales. Initializing with pre-trained weights from Huggingface\footnote{https://huggingface.co/}, we train the encoders with a fully connected layer added at the end of each encoder for NER prediction.
We also provide a comparison with XLM-R~\cite{conneau-etal-2020-unsupervised} in Appendix \ref{sec:appendix-xlmr}, with different set of languages for \textbf{Case 1}, \textbf{Case 2}, and \textbf{Case 3}.

\begin{figure}[t]
    \centering
    \includegraphics[width=\columnwidth]{
    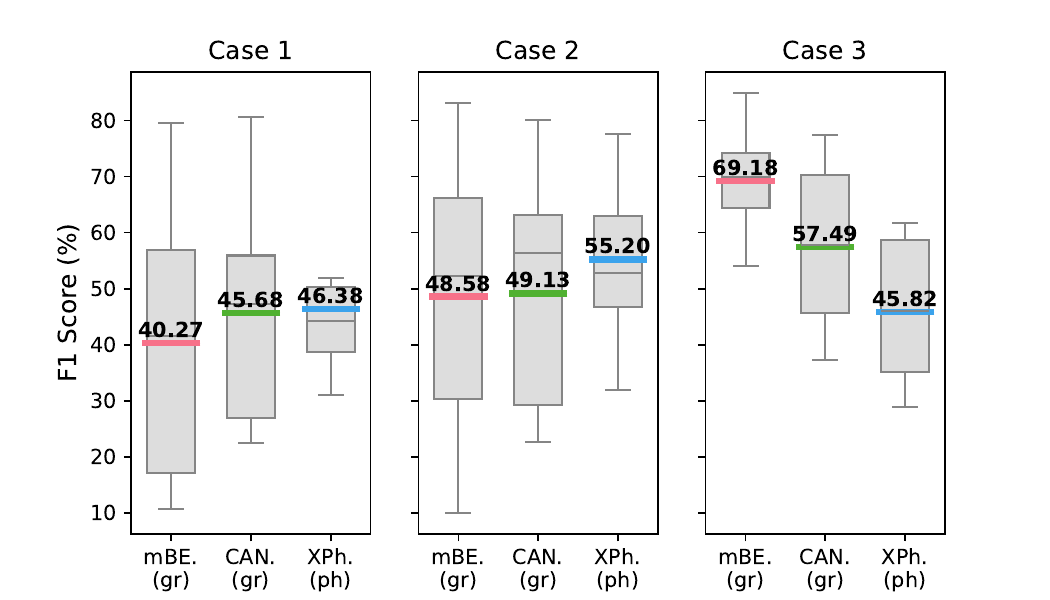
    }
    \caption{Distribution of F1 scores for each language set. X-axis shows each model using their first three letters, with `(gr)' and `(ph)' indicating their input forms (graphemes and phonemes, respectively). Colored horizontal lines and the numbers above show the average F1 scores for each model.}
    \label{fig:unseen-case-wise}
\end{figure}

\section{Results}

\begin{table*}[thb!]
\centering
\resizebox{\textwidth}{!}{
\begin{tabular}{c|c|cccccccc|cc}
\toprule
% CASE 1
\multirow{2}{*}{Input} & \multirow{2}{*}{Model} & \multicolumn{8}{c|}{Languages} & \multirow{2}{*}{AVG} & \multirow{2}{*}{STD} \\
\cmidrule{3-10}
&&sin & som & mri & quy & uig & aii & kin & ilo & & \\ \midrule
grapheme & mBERT &  10.71 & 44.76 & 38.48 & 55.07 & 18.70 & 12.58 & 62.37 & 79.51 & 40.27 & 25.00  \\
grapheme & CANINE & 26.31 & 43.35 & 51.30 & 59.48 & 27.19 & 22.38 & 54.74 & 80.70 & 45.68 & 19.99 \\
\rowcolor{LightBlue}phoneme (ours) & XPhoneBERT & \textbf{43.61} & 38.91 & 38.07 & 51.90 & \textbf{44.82} & \textbf{31.03} & 49.67 & 73.05 & \textbf{46.38} & \textbf{12.67} \\
\midrule
\bottomrule
\end{tabular}
}
\caption{\label{tab:main-table-case1}
Zero-shot performance in F1 scores (\%) on unseen languages (\textbf{Case 1}) using different models and input types.
}
\end{table*}

\subsection{Zero-Shot NER on Seen Languages}

Figure \ref{fig:unseen-case-wise} illustrates zero-shot performance of each model for each language set (\textbf{Case 1}, \textbf{Case 2}, and \textbf{Case 3}). Results on \textbf{Case 2} and \textbf{Case 3} align with our expectation, with languages seen during pre-training achieving better scores with the model. For the 12 languages in \textbf{Case 2}, XPhoneBERT, which was pre-trained on these languages, shows an average F1 score of 55.20\%, outperforming mBERT and CANINE by 6.62\% and 6.07\%, respectively. Languages of \textbf{Case 3} also performs better with models that were pre-trained on these languages. Specifically, mBERT achieves high scores for pre-trained languages, with average F1 score of 69.18\%, indicating its strong ability to generalize across seen languages. F1 scores for all models and languages are shown in Table \ref{tab:main-table-all} of Appendix.

\subsection{Zero-Shot NER on Unseen Languages}
Given the performance bias towards seen languages, we investigate the effect of using phonemes with languages that were not seen by any model---languages from \textbf{Case 1}. This ensures a fair comparison for low-resource languages, since extremely low-resource languages are often not included in the pre-training stage of language models. As shown in Table \ref{tab:main-table-case1}, the phoneme-based model demonstrates the best overall performance, achieving the highest scores on 3 out of 8 languages by a significant margin. Furthermore, the phoneme-based model exhibits the most stable performance across unseen languages, with the lowest standard deviation in scores.

\begin{figure}
    \centering
    \includegraphics[width=\columnwidth]{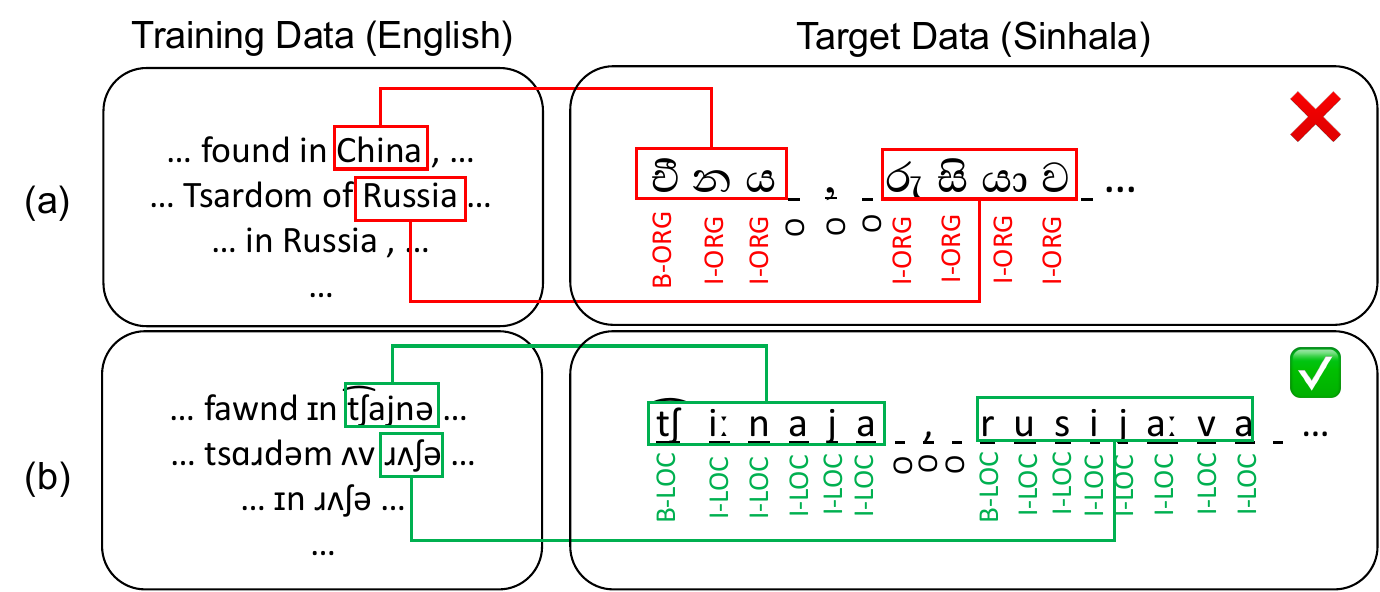}
    \caption{NER results on the target language (Sinhala) produced by each model trained on English data: (a) CANINE (b) XPhoneBERT.}
    \label{fig:qualitative-sin}
\end{figure}

Figure \ref{fig:qualitative-sin} shows a qualitative result of zero-shot inference on Sinhala, a language that is not in the pre-training data of any model. While the character-based model (a) fails to generalize to the language with different writing system, the phoneme-based model (b) successfully predicts the named entity tags due to the similar pronunciation of ``China'' and ``Russia'' across the languages.
These results indicate the robustness provided by phonemic representations, validating our hypothesis about the advantages they convey in NER tasks.

\begin{figure}[]
\vspace{-1em}
    \centering
    \includegraphics[width=\columnwidth]{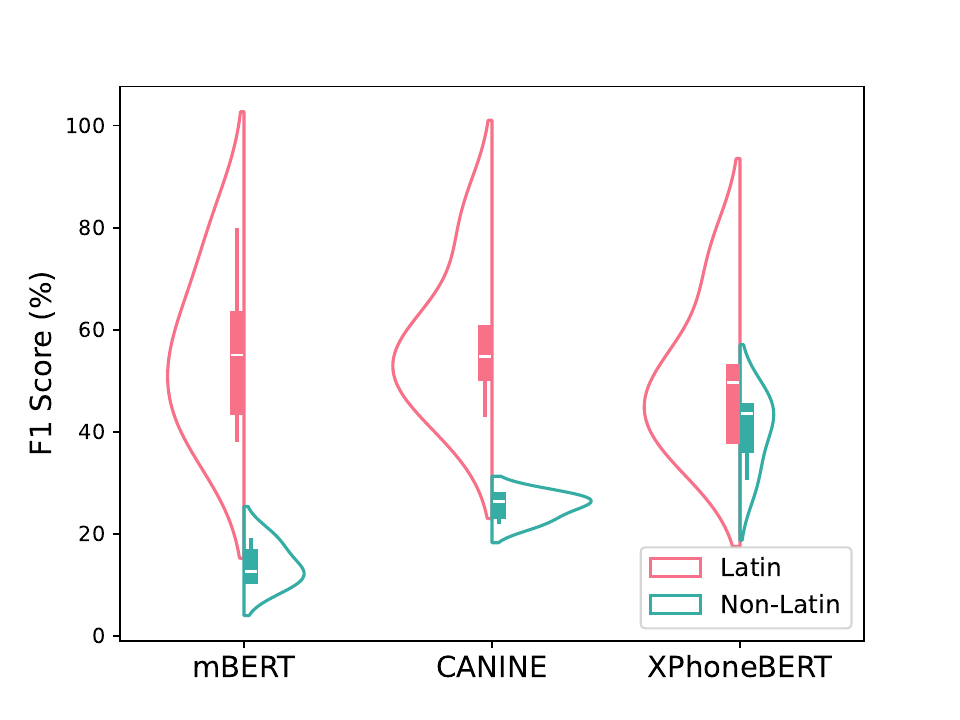}
    \caption{Performance distribution of each model on languages using Latin and non-Latin scripts from unseen languages.}
    \label{fig:case1-script}
\end{figure}

\subsection{Robustness Across Writing Systems}

One of the important advantages of using phonemic representations for named entity recognition is that it allows use of IPA. Using IPA for multilingual tasks provides a unified notation system. Observing the significant performance drop of mBERT on unseen low-resource languages (Figure \ref{fig:unseen-case-wise}), we consider this gap is largely attributed to the different writing systems of languages. Figure \ref{fig:case1-script} shows the distribution of F1 scores of each model on languages using Latin and non-Latin scripts from \textbf{Case 1}. mBERT, which performs the strongest on seen languages, exhibits the largest performance discrepancy between languages that use Latin and non-Latin scripts when evaluated on unseen languages. This highlights the limitation of the grapheme-based model, as it depends on the specific scripts. 

On the other hand, the phoneme-based model---XPhoneBERT---demonstrates the most consistent performance over different unseen languages with little performance gap between Latin and non-Latin scripts. This suggests that taking advantage of phonemes with its unified notation system allows for better generalization on extremely low-resource languages.

\section{Conclusion}
This paper presents the novel method of employing phonemes for identifying named entities for low-resource languages in zero-shot environments. 

Our experiments compared the results of phoneme-based models with grapheme-based models in a strict zero-shot setting, and have shown that phonemes exhibit the best performance over low-resource languages unseen by all models. The results particularly demonstrate robustness towards non-Latin scripts, which is crucial in context of multilingual NER since languages are written in diverse writing systems.

\section{Limitations}

One limitation is that we examined only the languages included in WikiANN dataset and G2P modules we employed, resulting in a comparison of a small number of completely unseen languages. Additionally, we used a limited number of baselines with models of restricted scales, making it difficult to ensure that the results would remain consistent if the models were more extensively tailored to the task.

Perhaps more concerning, the performance achieved by these approaches is not sufficient for production use. While this is probably to be expected of zero-shot approaches, it demonstrates how much work is left before these approaches have practical utility.

\section{Ethics Statement}
In this work, we use WikiANN~\cite{pan-etal-2017-cross} which is publicly available dataset to train various models with different languages. The WikiANN authors already grappled with many of the ethical issues involved in the curation and annotation of this resource. We did not find any outstanding ethical concerns, including violent or offensive content, though there are likely strong biases in the named entities represented in the data. We used the dataset as consistent with the intended use.
Nevertheless, we need to emphasize that, considering the characteristic of NER task, the dataset may contain personal information such as a specific person's real name or actual company name. We do not believe that this affects our result and the code and data distributed with our paper do not include any sensitive data of this kind.

\section*{Acknowledgments}
This work was supported by Institute of Information \& communications Technology Planning \& Evaluation (IITP) grant funded by the Korea government(MSIT) (RS-2022-00143911,AI Excellence Global Innovative Leader Education Program)(90\%) and ICT Creative Consilience Program through the Institute of Information \& Communications Technology Planning \& Evaluation(IITP) grant funded by the Korea government(MSIT)(RS-2020-II201819)(10\%).

\bibliography{acl_latex}

\appendix

\section{Implementation Details} \label{sec:appendix-implementation_details} We ran training on English subset of WikiANN dataset for 10 epochs, with learning rate of 1e-5, weight decay 0.01, batch size 128, and warmup ratio 0.025 on 1 NVIDIA RTX A5000 GPU. We set the maximum sequence length of the input 128 for all the models. We experimented with models of BERT-base scale: mBERT with 177M parameters, CANINE-C with 132M, and XPhoneBERT with 87M.

\section{Quantitative Results of Case 2 and Case 3}
We present the quantitative result of all three cases in Table \ref{tab:main-table-all}. The method using phoneme representation outperforms in Case 1 and Case 2 in terms of average F1 score and demonstrates more stable results with a lower standard deviation.

\begin{table*}
\centering
\resizebox{\textwidth}{!}{
\begin{tabular}{c|ccccccccccccccccc}
\toprule
% CASE 1
Case & Input & Model & \multicolumn{13}{c}{Languages} & AVG & STD \\
\midrule
\multirow{4}{*}{CASE 1}&&&sin & som & mri & quy & uig & aii & kin & ilo & & & & & & & \\ \cmidrule{2-18}
 & grapheme & mBERT & 10.71 & \textbf{44.76} & 38.48 & 55.07 & 18.70 & 12.58 & \textbf{62.37} & 79.51 & & & & & & 40.27 & 25.00 \\
& grapheme & CANINE & 26.31 & 43.35 & \textbf{51.30} & \textbf{59.48} & 27.19 & 22.38 & 54.74 & \textbf{80.70} & & & & & & 45.68 & 19.99  \\
\rowcolor{LightBlue}\cellcolor{white}&phoneme (ours) & XPhoneBERT & \textbf{43.61} & 38.91 & 38.07 & 51.90 & \textbf{44.82} & \textbf{31.03} & 49.67 & 73.05 & & & & & & \textbf{46.38} & \textbf{12.67}  \\
\midrule
% CASE 2
\multirow{4}{*}{CASE 2} & & & epo & khm & tuk & amh & mlt & ori & san & ina & grn & bel & kur & snd & &  &  \\
\cmidrule{2-18}
& grapheme & mBERT & 71.31 & 16.12 & \textbf{64.52} & 11.90 & \textbf{63.83} & 9.96 & 48.73 & \textbf{73.89} & 50.44 & \textbf{83.12} & 54.16 & 35.02 & & 48.58 & 25.13 \\
& grapheme & CANINE & 68.19 & 27.33 & 58.07 & 22.65 & 61.58 & 33.53 & 26.79 & 68.78 & \textbf{55.37} & 80.07 & \textbf{57.33} & 29.87 & & 49.13 & 19.86 \\
\rowcolor{LightBlue}\cellcolor{white}&phoneme (ours) & XPhoneBERT & \textbf{75.26} & \textbf{31.86} & 61.17 & \textbf{44.85} & 52.58 & \textbf{40.73} & \textbf{59.42} & 68.68 & 49.95 & 77.61 & 52.95 & \textbf{47.28} & & \textbf{55.20} & \textbf{13.83}  \\
\midrule
% CASE 3
\multirow{4}{*}{CASE 3}& & & tgk & yor & mar & jav & urd & msa & ceb & hrv & mal & tel & uzb & pan & kir &  &  \\
\cmidrule{2-18}
 & grapheme & mBERT & \textbf{74.10} & \textbf{56.60} & \textbf{74.30} & \textbf{73.59} & \textbf{57.09} & 74.98 & 64.44 & \textbf{84.93} & \textbf{69.94} & \textbf{67.24} & \textbf{80.04} & \textbf{53.98} & \textbf{68.14} & \textbf{69.18} & \textbf{9.28} \\
& grapheme & CANINE & 62.12 & 51.15 & 44.28 & 61.11 & 42.41 & \textbf{76.82} & \textbf{70.36} & 77.51 & 48.29 & 37.29 & 72.54 & 45.74 & 57.73 & 57.49 & 13.77 \\
\rowcolor{LightBlue}\cellcolor{white}&phoneme (ours) & XPhoneBERT & 48.93 & 50.87 & 35.12 & 45.98 & 33.37 & 61.76 & 58.72 & 58.76 & 32.52 & 28.93 & 60.92 & 43.85 & 35.95 & 45.82 & 11.85 \\
\bottomrule
\end{tabular}
}
\caption{\label{tab:main-table-all}
Zero-shot F1 score (\%) result in \textbf{Case 1}, \textbf{2}, and \textbf{3}.
\vspace{-1em}
}
\end{table*}

\begin{figure}[!h]
    \centering
    \includegraphics[width=\columnwidth]{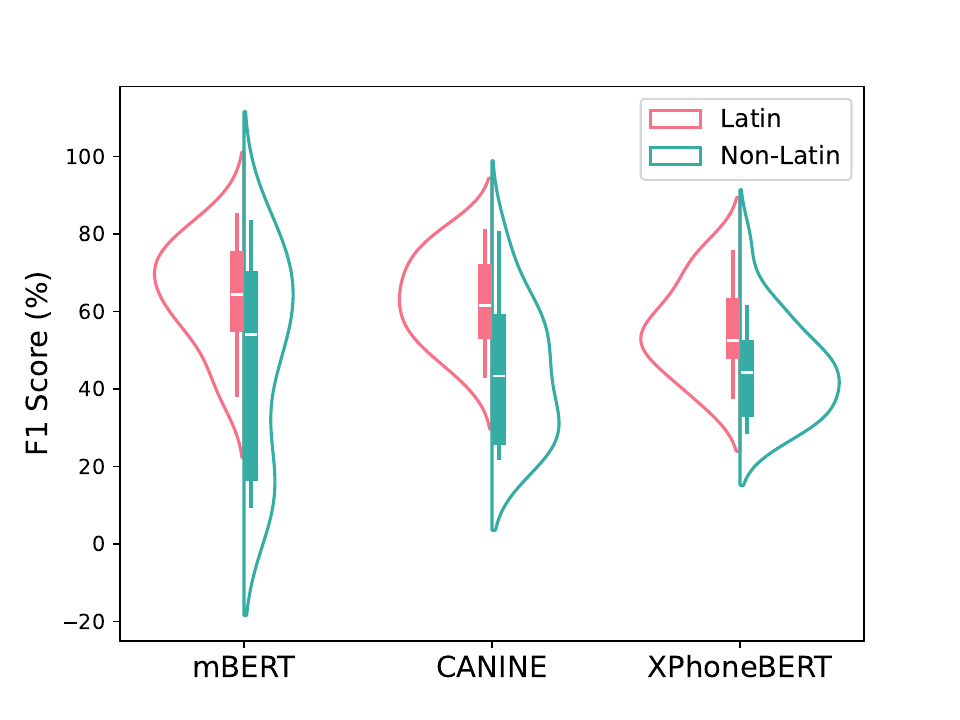}
    \vspace{-2em}
    \caption{Performance distribution of each model on languages using Latin and non-Latin scripts.}
    \label{fig:all-latin-nonlatin}
\end{figure}

\section{Comparison of Latin and Non-Latin Languages}
Figure \ref{fig:all-latin-nonlatin} shows the performance distribution of zero-shot NER on all languages, where the performance distribution of languages using Latin and non-Latin scripts are visualized separately.
Compared to mBERT and CANINE that exhibit significant performance gaps between Latin and non-Latin scripts, XPhoneBERT shows little difference in performance distribution.

\begin{table*}[!h]
\centering
\resizebox{\textwidth}{!}{
\begin{tabular}{c|ccccccccccccccc}
\toprule
% CASE 1
Case & Input & Model & \multicolumn{11}{c}{Languages} & AVG & STD \\
\midrule
\multirow{3}{*}{CASE 1}&&& mri & quy & aii & kin & ilo & tgk & yor & ceb & & & & & \\ \cmidrule{2-16}
 & grapheme & XLM-R & 29.93 & 65.09 & 14.58 & 49.05 & 71.29 & 40.96 & 55.01 & 64.31 & & & & 48.78 & 19.42 \\
\rowcolor{LightBlue}\cellcolor{white}&phoneme (ours) & XPhoneBERT & 38.07 & 51.90 & 31.03 & 49.67 & 73.05 & 48.93 & 50.87 & 58.72 & & & & \textbf{50.28} & \textbf{12.62}  \\
\midrule
% CASE 2
\multirow{3}{*}{CASE 2} & & & tuk & mlt & ina& grn & & & & & & & & &  \\
\cmidrule{2-16}
& grapheme & XLM-R & 53.63 & 58.81 & 72.55 & 44.80 &  & & & & &  &  & 57.45 & 11.61 \\
\rowcolor{LightBlue}\cellcolor{white}&phoneme (ours) & XPhoneBERT & 61.17 & 52.58 & 68.68 & 49.95 & & &  & & & & & \textbf{58.10} & \textbf{8.53} \\
\midrule
% CASE 3
\multirow{3}{*}{CASE 3}& & & epo & khm & amh & ori & san & bel & kur & snd & sin & som & uig &  &  \\
\cmidrule{2-16}
 & grapheme & XLM-R & 73.82 & 48.85 & 64.89 & 54.93 & 59.92 & 82.93 & 75.61 & 42.90 & 64.91 & 55.52 & 61.74 & \textbf{62.37} & \textbf{11.88} \\
\rowcolor{LightBlue}\cellcolor{white}&phoneme (ours) & XPhoneBERT & 75.26 & 31.86 & 44.85 & 40.73 & 59.42 & 77.61 & 52.95 & 47.28 & 43.61 & 38.91 & 44.82 & 50.66 & 14.60 \\
\bottomrule
\end{tabular}
}
\caption{\label{tab:xlm-xphonebert}
Zero-shot F1 score (\%) result in \textbf{Case 1}, \textbf{2}, and \textbf{3} languages specifically selected for XLM-R and XPhoneBERT.
}
\end{table*}

\section{Comparison with XLM-R}
\label{sec:appendix-xlmr}
Here we provide additional results of zero-shot inference of XLM-R(base) \cite{conneau-etal-2020-unsupervised} and XPhoneBERT \cite{nguyen2023xphonebert}. XLM-R shares the training objective with XPhoneBERT, removing the variance that comes from the difference in pre-training objectives. 
Since XLM-R is trained on different set of languages, we listed up another sets of languages that corresponds to \textbf{Case1}, \textbf{Case2}, and \textbf{Case3}, which refers to the set of languages not trained on both models, languages trained on XPhoneBERT not on XLM-R, and languages trained on XLM-R not on XPhoneBERT, respectively.

As shown in Table \ref{tab:xlm-xphonebert}, the result for \textbf{Case1} is consistent with our findings in the main experiments, exhibiting relatively stable performance with XPhoneBERT (higher average and lower standard deviation). Also, languages that are written in non-Latin scripts perform significantly better with XPhoneBERT, which aligns with our analysis in the main text as well.

\section{Language Codes}
In Table \ref{tab:lang_code}, we have listed ISO 639-3 language codes of all languages used in the experiments.

\begin{table}
\centering
\resizebox{0.7\columnwidth}{!}{
\begin{tabular}{ll}
\toprule
Language & ISO 639-3 \\
\midrule
Amharic & amh \\
Assyrian Neo-Aramaic & aii\\ 
Ayacucho quechua & quy\\ 
Cebuano & ceb\\
Croatian & hrv\\
English & eng\\
Esperanto & epo\\
Ilocano & ilo\\
Javanese & jav\\
Khmer & khm\\
Kinyarwanda & kin \\
Korean & kor \\
Kyrgyz & kir \\
Malay & msa \\
Malayalam & mal \\
Maltese & mlt \\
Maori & mri \\
Marathi & mar \\
Punjabi & pan \\
Sinhala & sin \\
Somali & som \\
Spanish & spa \\
Tajik & tgk \\
Telugu & tel \\
Turkmen & tuk \\
Urdu & urd \\
Uyghur & uig \\
Uzbek & uzb \\
Yoruba & yor \\
\bottomrule
\end{tabular}}
\caption{\label{tab:lang_code}
Language codes for all languages used in the experiments.
}
\end{table}

\section{Benchmark and License}
In Table \ref{tab:dataset_stats}, we provide the datasets, their statistics, and license. We also used CharsiuG2P~\cite{Zhu2022ByT5MF} toolkit for transliteration, which is under MIT license.

\begin{table}[]
    \centering
    \scriptsize
    \begin{tabular}{ccccccc}
    \toprule
         Dataset & Lang. & Script & Train & Dev & Test & License \\ \midrule
         \multirow{34}{*}{WikiANN} & \texttt{eng}& Latn & 20k & 10k & 10k & \multirow{34}{*}{ODC-BY} \\
                                & \texttt{sin}& Sinh & 100 & 100 & 100 & \\
                                & \texttt{som}& Latn & 100 & 100 & 100 & \\
                                & \texttt{mri}& Latn & 100 & 100 & 100 & \\
                                & \texttt{quy}& Latn & 100 & 100 & 100 & \\
                                & \texttt{uig}& Arab & 100 & 100 & 100 & \\
                                & \texttt{aii}& Syrc & 100 & 100 & 100 & \\
                                & \texttt{kin}& Latn & 100 & 100 & 100 & \\
                                & \texttt{ilo}& Latn & 100 & 100 & 100 & \\
                                & \texttt{epo}& Latn & 15k & 10k & 10k & \\
                                & \texttt{khm}& Khmr & 100 & 100 & 100 & \\
                                & \texttt{tuk}& Latn & 100 & 100 & 100 & \\
                                & \texttt{amh}& Ethi & 100 & 100 & 100 & \\
                                & \texttt{mlt}& Latn & 100 & 100 & 100 & \\
                                & \texttt{ori}& Orya & 100 & 100 & 100 & \\
                                & \texttt{san}& Deva & 100 & 100 & 100 &  \\
                                & \texttt{ina}& Latn & 100 & 100 & 100 &  \\
                                & \texttt{grn}& Latn & 100 & 100 & 100 & \\
                                & \texttt{bel}& Cyrl & 15k & 1k & 1k & \\
                                & \texttt{kur}& Latn & 100 & 100 & 100 & \\
                                & \texttt{snd}& Arab & 100 & 100 & 100 & \\
                                & \texttt{tgk}& Cyrl & 100 & 100 & 100 & \\
                                & \texttt{yor}& Latn & 100 & 100 & 100 &  \\
                                & \texttt{mar}& Deva & 5k & 1k & 1k &  \\
                                & \texttt{jav}& Latn & 100 & 100 & 100 & \\
                                & \texttt{urd}& Arab & 20k & 1k & 1k & \\
                                & \texttt{msa}& Latn & 20k & 1k & 1k & \\
                                & \texttt{ceb}& Latn & 100 & 100 & 100 & \\
                                & \texttt{hrv}& Latn & 20k & 10k & 10k & \\
                                & \texttt{mal}& Mlym & 10k & 1k & 1k &  \\
                                & \texttt{tel}& Telu & 1k & 1k & 1k &  \\
                                & \texttt{uzb}& Cyrl & 1k & 1k & 1k & \\
                                & \texttt{pan}& Guru & 100 & 100 & 100 & \\
                                & \texttt{kir}& Latn & 100 & 100 & 100 & \\
         \bottomrule
  \end{tabular}
  \caption{Statistics and license types for the dataset. The table lists the script, number of examples in the training, development, and testing sets for languages in the WikiANN dataset. The dataset is strictly used within the bounds of these licenses.}
      \label{tab:dataset_stats}
\end{table}

\end{document}